\documentclass[10pt,twocolumn,letterpaper]{article}
\pdfoutput=1
\usepackage{cvpr}
\usepackage{multirow}
\usepackage{times}
\usepackage{epsfig}

\usepackage{graphicx}
\usepackage{amsmath}
\usepackage{amssymb}
\usepackage{subcaption}
\usepackage{pifont}
\usepackage{xcolor}
\usepackage{makecell}
\usepackage{color,soul}

\usepackage{booktabs}

\usepackage[pagebackref=true,breaklinks=true,letterpaper=true,colorlinks,bookmarks=false,citecolor=green]{hyperref}
\definecolor{cerulean}{rgb}{0.0, 0.48, 0.65}
\hypersetup{
     citecolor = cerulean,
     }

\usepackage{amsmath,amsfonts,bm}

\def\eqref#1{equation~\ref{#1}}

\def\1{\bm{1}}

\DeclareMathAlphabet{\mathsfit}{\encodingdefault}{\sfdefault}{m}{sl}
\SetMathAlphabet{\mathsfit}{bold}{\encodingdefault}{\sfdefault}{bx}{n}

\cvprfinalcopy 

\def\cvprPaperID{5827} 

\begin{document}

\title{Optimized Generic Feature Learning for Few-shot Classification across Domains}

\author{Tonmoy Saikia\thanks{Work mainly done while interning at Google.}\\
 University of Freiburg\\
{\tt\small saikiat@cs.uni-freiburg.de}
\and
Thomas Brox\\
University of Freiburg\\
{\tt\small brox@cs.uni-freiburg.de}
\and
Cordelia Schmid\\
Google Research\\
{\tt\small cordelias@google.com}
}

\maketitle

\newcommand{\tct}[1]{\textcolor{blue}{Tonmoy: #1}} 
\newcommand{\draft}[1]{\hl{#1}} 
\newcommand{\tcc}[1]{\textcolor{red}{#1}} 
\newcommand{\tcb}[1]{\textcolor{olive}{\textbf{Thomas: #1}}} 

\newcommand{\cmark}{\ding{51}}%
\newcommand{\xmark}{\ding{55}}%

\newcommand{\rot}[1]{\rotatebox[origin=c]{90}{\parbox[c]{1cm}{\centering #1}}}

 \begin{abstract}
 To learn models or features that generalize across tasks and domains is one of the grand goals of machine learning.
 In this paper, we propose to use cross-domain, cross-task data as validation objective for hyper-parameter optimization (HPO) to improve on this goal.
 Given a rich enough search space, optimization of hyper-parameters learn features that maximize validation performance and, due to the objective, generalize across tasks and domains.
 We demonstrate the effectiveness of this strategy on few-shot image classification within and across domains.
 The learned features outperform all previous few-shot and meta-learning approaches.
 \end{abstract}


\section{Introduction}

Generalization to unseen samples distinguishes machine learning from traditional data mining. For machine learning, it is not enough to store sufficiently many examples and to access them efficiently, but the examples are supposed to yield a generic model, which makes reasonable decisions (far) beyond the observed training samples. Pushing the limits of generalization is probably the most important goal of machine learning. 

To measure generalization, typically a dataset is split into a training and a test set. With no overlap between these sets, the test error indicates generalization to unseen samples. However, data in the training and test set are typically very similar. 
More challenging test schemes are cross-dataset testing and few-shot learning. In cross-dataset testing, the test set is from an independent dataset, which involves a considerable domain shift. 
In few-shot learning, a trained model is tested on unseen classes. Only few (sometimes one) samples of the new classes are made available to learn their decision function. This is only possible, if the features from the pre-trained model are rich enough to provide the discriminative information for the new classes: the features must generalize to new tasks. 

Deep networks exhibit far more parameters than there are training samples. Without appropriate regularization, they would learn the training samples and not generalize at all. However, we all know that deep networks \emph{do} generalize.
This is partially due to explicit regularization, such as weight decay, dropout, data augmentation, or ensembles, and also due to implicit regularization imposed by the optimization settings and network architecture~\cite{gen_bounds,novak2018sensitivity,zhang_rethinking_gen}. Although the research community has developed good practices for all the many design choices, it is difficult to find the right balance, since there are mutual dependencies between parameters~\cite{fanova}. Even worse, the right balance depends much on the training set the network draws its information from: training on ImageNet can require very different regularization than training on a smaller, less diverse dataset. 

In this paper, we propose rigorous hyperparameter optimization (HPO) to automatically find the optimal balance between a set of factors that influence regularization. The use of HPO for improving the test performance of deep networks is very common \cite{autoaugment,bohb,pbt_augment,pbt,hyperband}. What distinguishes this paper from previous works, is that we analyze the suitability of HPO for improving generalization rather than optimal test performance on a particular benchmark dataset. To this end, we choose the validation objective to match what we want: generalization. We investigate two levels of generic feature learning: (1) within-domain generalization in few-shot learning, and (2) cross-domain generalization. Few-shot learning often does not work well in unseen domains~\cite{closerlook,ensembles}. 
Thus, in case (2) we investigate if HPO can improve cross-domain performance when cross-domain tasks are used for optimization. 

Our investigation gave us several scientific insights. 
\textbf{(1)}~We show that HPO boosts few-shot performance clearly beyond the state-of-the-art, even with a rather small search space. This confirms that manually finding the right balance in the set of options that affect regularization is extremely hard. \textbf{(2)} Cross-domain HPO further improves few-shot performance in the target domain. \textbf{(3)} The search confirms some common best practice choices and discovers some new trends regarding the batch size, the choice of the optimizer, and the way to use data augmentation.

\begin{figure*}[t]
\begin{center}
\includegraphics[width=0.85\linewidth]{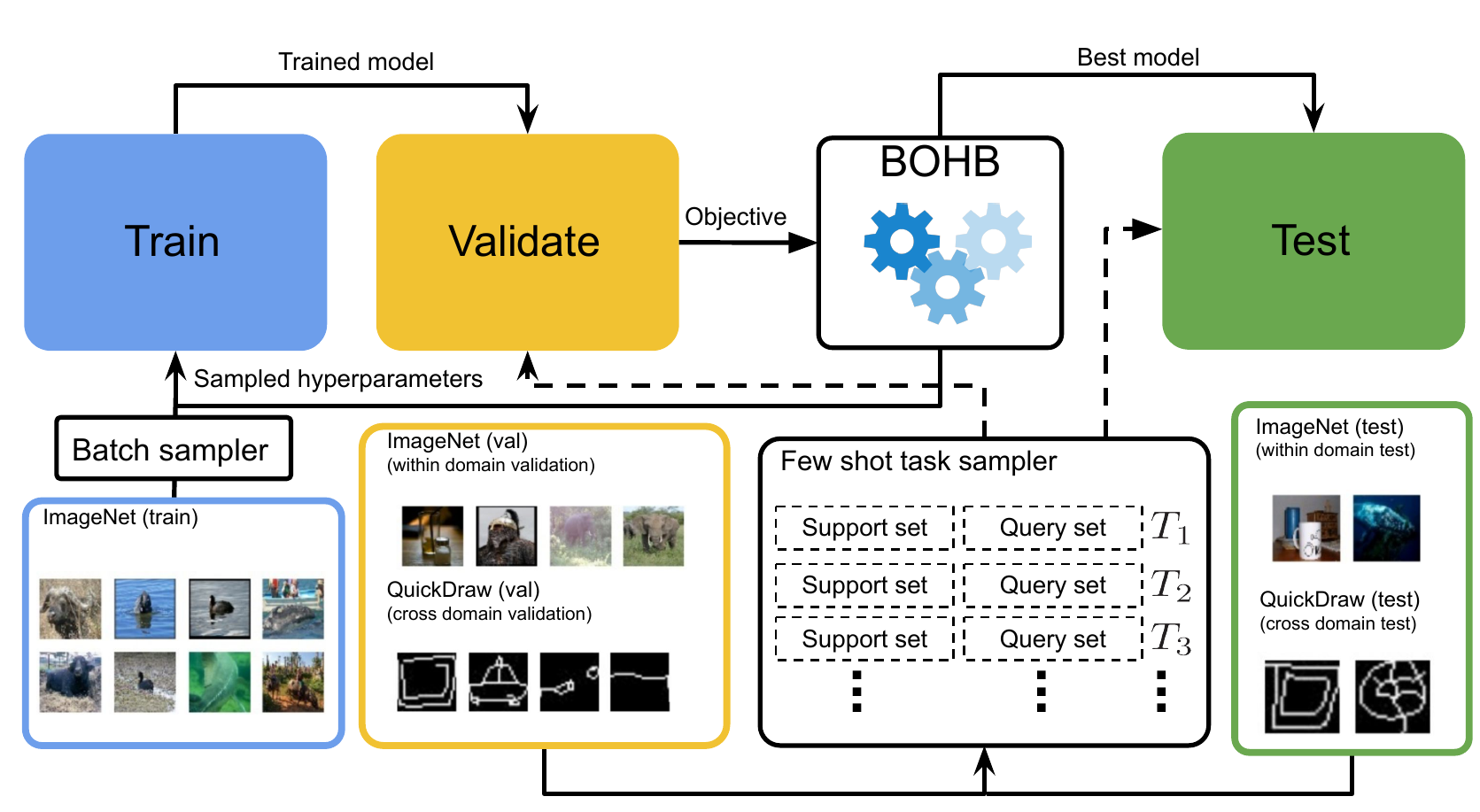}
\caption{Hyperparameter optimization (train, validation and test stages). Multiple feature extractors are learned on the training set with hyperparameters sampled by BOHB. The validation objective is the average accuracy over 
multiple few-shot tasks sampled from the validation split.
A few-shot task $T_i$ consists of a support set  and a query set. For each few-shot task a classifier is learned based on the support set and the feature extractor from the training stage. The query set is used for evaluation, i.e., to measure classification accuracy. The best model is then evaluated on the test set using the same protocol as during validation. Few-shot tasks may be from a domain different from that of the training set.}
\label{fig:pipeline}
\end{center}

\end{figure*}

\section{Related work}
\label{relatedwork}

\textbf{Feature learning.}  A standard approach to obtain a common feature representation is to train a classification network on ImageNet. These features generalize surprisingly well and are commonly used for finetuning on target domains or tasks~\cite{girshick2015fast, girshick_richfeature,  fcseg, overfeat}. Studies have shown that data augmentation~\cite{autoaugment,randaugment,fastautoaugment} and regularization techniques~\cite{shakeshake,dropblock} can help learn feature representations which generalize better. Unsupervised approaches use proxy tasks and loss functions to learn good feature representations. Examples of proxy tasks could be recovering images under corruption~\cite{context_encoders,denoising_auto} or image patch orderings~\cite{patch_orderings,jigsaw}. Loss functions focus on contrastive~\cite{replearn,wu2018unsupervised} or adverserial learning~\cite{advfeatures,advfeatures_large}.
In most cases, unsupervised feature learning is not competitive with features obtained from ImageNet. 

\textbf{Few-shot learning.} Few-shot learning approaches can primarily be categorized into optimization~\cite{maml,opt_model,leo,metadataset} and metric learning approaches~\cite{koch2015siamese,relation_net,matchingnet}.
Optimization approaches are designed with the goal to adapt quickly to new few-shot tasks.
Ravi and Larochelle~\cite{opt_model} used an LSTM to learn an update rule for a good initialization for a model's parameters, such that it performs well given a small training set on a new classification task. 
Finn \etal~\cite{maml} proposed MAML which uses gradient updates obtained from target tasks (unseen during training) to learn better weight initialization to prevent over-fitting. However, methods such as MAML have been shown to work poorly on larger datasets~\cite{metadataset}. State-of-the-art methods~\cite{metaoptnet,tadam,protonet} often rely on metric learning~\cite{dhillon2019baseline}. Typically such methods meta-learn an embedding over multiple meta-training  tasks~\cite{matchingnet} which are then used to classify query examples based on some distance metric. For instance, Vinyals~\etal~\cite{matchingnet} proposed matching networks, which label query set examples as a linear combination of the support set labels. Examples in the support set with embedding representations closer to the query set example are assigned a higher weight. Snell~\etal~\cite{protonet} assume that there exists an embedding space where samples of the same class cluster around a prototype. Their method builds class prototypes by taking the mean of the support set embedding for each example. Query set examples are then assigned class labels based on their closeness to class prototypes. 

Recently, it has been shown that the standard feature learning approach, of first training a backbone network on a large number of base classes and learning a few-shot classifier on top, is competitive~\cite{closerlook,ensembles,metadataset}. In this work, we use standard feature learning.

\textbf{Hyperparameter optimization.}
Bayesian optimization is the most common approach for hyperparameter search and has been used to obtain state-of-the-art results on CIFAR-10~\cite{snoek}. Standard Bayesian optimization does not scale well to tasks with longer training time or large hyperparameter search spaces. In such cases, methods that run mostly random search like Hyperband~\cite{hyperband} are often more efficient. BOHB~\cite{bohb} adds Bayesian optimization to Hyperband to learn from previously sampled hyperparameter configurations. In computer vision, BOHB was used to optimize training a disparity network~\cite{autodispnet}. 
We use BOHB for hyperparameter optimization targeted towards optimal features for few-shot learning tasks.



\section{Optimized feature learning}

\label{sec:methods}

To optimize generalization via hyperparameter optimization (HPO), we must define an appropriate search space, an informative validation objective, and the dataset splits on which we optimize. An overview of the whole optimization pipeline with the different data splits is shown in Figure~\ref{fig:pipeline}.

\subsection{Hyperparameter search space}
\label{sec:search_space}

The definition of the search space is decisive for the success of HPO. With a naive way of thinking, one would make the search space as large as possible to explore all possibilities. However, this will make HPO terribly inefficient, and the search will not find optimal values even after many GPU years. 
Thus, some informed choice of interesting hyperparameters has to be made. We experiment with two search spaces, which are motivated by best practice insights of previous works~\cite{small_batch_training,sgd_generalization,zhang_rethinking_gen} and by scientific questions that we want to get answered in this work.  

\textbf{Search space S1 (Optimization).} In this smaller search space we consider five training parameters, which are relevant for regularization: the choice of optimizer, the batch size, the amount of $L_2$ regularization, the initial learning rate, and its decay frequency, i.e., the number of mini-batch updates until the learning rate is decreased. The learning rate is decayed to $0$ following a cosine schedule~\cite{cosine_decay}.

\textbf{Search space S2 (Optimization \& Augmentation).} Since data augmentation plays an important role for regularization, we define a second, larger search space that includes data augmentation on top of S1. Search spaces on data augmentation can quickly become very large. AutoAugment~\cite{autoaugment} searched for optimal augmentation policies, i.e., sequences of augmentation operations, using reinforcement learning. A huge search space tends to yield sub-optimal results after finite compute time. Indeed, in a follow-up work Cubuk~\etal~\cite{randaugment} claimed that the full policy search was unnecessary, as randomly choosing $N$ transformations from the overall set of transformations performs on-par.
This leaves the question on the table: is this a) because it is not critical to tune exact magnitudes for each augmentation operation as long as there is a diverse set of operations, or b) because the search with reinforcement learning found a sub-optimal policy? 

We shed some more light on this using a smaller search space than in AutoAugment, i.e.,  
 a subset of 10 operations from AutoAugment (rotate, posterize, solarize, color, contrast, brightness, sharpness, shear, translate, cutout~\cite{cutout}) and only a single magnitude parameter for each of them. In particular, we sample the transformation magnitude from a Gaussian distribution clipped at $0$ and optimize the standard deviation parameter (search range described in Section~\ref{sec:exp_setup}). We allow augmentation operations to be applied at random and optimize the number of operations $N_{ops}$ applied at a time, which yields a single additional parameter. 

\subsection{Validation objective}
\label{sec:validation}
Apart from the training objective, which is optimized on the training set, HPO requires an additional validation objective on a held out validation set. 
We propose to use the accuracy of few-shot classification as validation objective to measure generalization performance. The overall scheme, as illustrated in Figure~\ref{fig:pipeline}, contains multiple training and optimization stages and different data subsets used for these. We describe it in detail in the following subsections.

\subsubsection{Training a classification network for feature extraction}
\label{subsubsec:trainfeatures}
The inner part of the optimization is training the weights $\theta$ of a normal classification network using the hyperparameter sample $h$ currently evaluated by HPO. We use the ResNet18 implementation from \cite{metadataset}. Each training run with hyperparameters $h$ yields a different network with weights $\theta_h$ and corresponding features $F_h$. 

\subsubsection{Sampling of few-shot learning tasks}
The features $F_h$ are validated on how well they perform as basis for a few-shot learning task. To this end, we evaluate the average result of multiple few-shot classification tasks, which are randomly sampled from the validation set. 

Each few-shot task $T_{i}$ consists of $n_i$ classes, with a few labeled examples ($k_i$ per class). For each task, a classifier is trained on these few labeled examples (\emph{support set})
and is evaluated on unseen examples of the same task (\emph{query set}). The number of classes in the task is commonly called \emph{ways} and the number of labelled examples is termed as \emph{shots}. We measure for each task $T_{i}$ the classification accuracy on its query set. 
Each task may have variable number of ways and shots. 

\subsubsection{Training the few-shot classifier for a task}
For classifying images in a few-shot task, we freeze the parameters $\theta$ of the feature extractor and train a classifier $C_{W_n}$, where $W_n \in \mathbb{R}^{d \times c_n}$ and $c_n$ represents the number of classes or ways, on the support set.
This classifier takes the feature embedding $F_h$ as input. 

We experiment with two types of classifiers: a linear classifier and a nearest centroid classifier~\cite{nearest_centroid}. 
The weight matrix ${W_n}$ for the linear classifier is trained using the cross entropy loss. 
The nearest centroid classifier~\cite{nearest_centroid} involves computing a class prototype by taking the average of embeddings from each class' support set examples. 
Each query set example is assigned the class, whose prototype is nearest according to the negative cosine similarity~\cite{dynamic_fewshot}. 

\subsubsection{Optimizing the hyperparameters with BOHB}
\label{sec:bohb}

The task of BOHB is to sample hyperparameters $h$ to train a feature extractor, $F_h$, such that the average accuracy over the few-shot tasks on the validation set is maximized. We emphasize that $h$ refers to the hyperparameters for training the feature extractor and not the weights $W_{n}$ of the few-shot classifiers. 

BOHB~\cite{bohb} combines the benefits of Bayesian optimization~\cite{Shahriari16}
and Hyperband~\cite{hyperband}.  Hyperband performs efficient random search by dynamically allocating more resources to promising configurations and stopping bad ones early. 
To improve over Hyperband, BOHB replaces its random sampling with model based sampling once there are enough samples to build a reliable model.

To improve efficiency, BOHB uses cheaper approximations $˜g(\cdot, b)$ of the validation objective $g(\cdot)$ , where $b$ refers to a budget configuration with $b \in [b_{min}, b_{max}]$, where the true validation objective is recovered with $b=b_{max}$. The budget usually refers to the number of mini-batch updates or number of epochs~\cite{bohb, autodispnet} but may also be other parameters such as image resolution or training time~\cite{nas_hp}. We use the number of mini-batch updates as  budget. Similar to Hyperband, BOHB repeatedly calls Successive Halving (SH)~\cite{shalving} to advance promising configurations evaluated on smaller budgets to larger ones. SH starts with a fixed number of configurations on the cheapest budget $b_{min}$ and retains the the best fraction ($\eta^{-1}$) for advancing to the next and more expensive budget. BOHB uses a multivariate kernel density estimator (KDE) to model the densities of the good and bad configurations and uses them to sample promising hyperparameter configurations. For more details we refer to the original paper~\cite{bohb}. 

\subsection{Testing the generalization performance}
At test time, we take the best model found by BOHB, and evaluate it on few-shot tasks sampled from the test set. Like the validation objective, we compute the test objective as the average classification accuracy over few-shot tasks. 
\section{Experiments and results}

With the experiments we are interested in answering the following scientific questions: 
\textbf{(1)} Does HPO improve generalization performance as measured by few-shot tasks?
\textbf{(2)}~Does HPO help transfer features to a target domain? 
\textbf{(3)} Can it yield features that generalize better to unseen domains? 
\textbf{(4)} How critical is the optimization of data augmentation parameters?
\textbf{(5)} How much do we gain by ensembling top performing models from HPO?

\subsection{Experimental setup}
\label{sec:exp_setup}

\textbf{Datasets.} We use mini-ImageNet~\cite{matchingnet} and datasets from the more challenging Meta-Dataset benchmark~\cite{metadataset}. 

The mini-ImageNet dataset is commonly used for few-shot learning. It consists of $600$ images per class. The training, validation and testing splits consist of $64$, $16$, and $20$ classes each.
Since all data is from the same dataset, it only enables evaluating cross-task, within-domain generalization, but not generalization across domains. 

To this end, we use the datasets from the Meta-Dataset benchmark. It is much larger than previous few-shot learning benchmarks and consists of multiple datasets of different data distributions. Also, it does not restrict few-shot tasks to have fixed ways and shots, thus representing a more realistic scenario. It consists of $10$ datasets from diverse domains: ILSVRC-2012~\cite{imagenet} (the ImageNet dataset, consisting of natural images with $1000$ categories), Omniglot~\cite{omniglot} (hand-written characters, $1623$ classes), Aircraft~\cite{aircraft} (dataset of aircraft images, 100 classes) , CUB-200-2011~\cite{birds} (dataset of Birds, 200 classes), Describable Textures~\cite{dtd} (different kinds of texture images with 43 categories), Quick Draw~\cite{quickdraw} (black and white sketches of 345 different categories), Fungi~\cite{fungi} (a large dataset of mushrooms with ~$1500$ categories), VGG Flower~\cite{vggflower} (dataset of flower images with $102$ categories), Traffic Signs~\cite{trafficsigns} (German traffic sign images with 43 classes) and MSCOCO (images collected from Flickr, $80$ classes). All datasets except Traffic signs and MSCOCO have a training, validation and test split  (proportioned roughly into $70\%$, $15\%$, $15\%$). The datasets Traffic Signs and MSCOCO are reserved for testing only. We note that there exist other ImageNet splits, such as TieredImageNet~\cite{tiered_imagenet}, therefore to avoid ambiguity we refer to Meta-Dataset's ImageNet version as ImageNet-GBM (\textbf{G}oogle \textbf{B}rain \textbf{M}ontreal). 

\begin{table}[h]
\begin{center}
\caption{Dataset groups (Meta-Dataset). $D1$ is used for training and evaluation whereas $D2$ and $D3$ are used for evaluations only.}
\label{tab:dspool}
\resizebox{0.65\linewidth}{!}{
\begin{tabular}{ll}
\hline
\textbf{Group}    &  \textbf{Datasets} \\
\hline
$D1$              & ImageNet-GBM  \\
$D2$              & \makecell[l]{Birds, Omniglot, QuickDraw,\\ DTD, Fungi, Aircraft, VGG Flower}               \\
$D3$              & MSCOCO, TrafficSign \\
\hline
\end{tabular}

}
\end{center}
\end{table}
We group the datasets in the Meta-Dataset benchmark into three subsets $D1$, $D2$, $D3$ as shown in Table \ref{tab:dspool}. A dataset in each subset has a training, validation, and test split. We always use only the respective split for the training, validation and testing stages. Testing on $D2$, for example, means using the test splits of the datasets in $D2$.
The splits of $D1$ (ImageNet-GBM) are used for training, validation, and testing, while the splits of $D2$ are used only for validation and testing. $D3$ is reserved for testing. 

\textbf{Evaluation details.} We compute the average accuracy over $600$ few-shot tasks sampled from a dataset. Few-shot tasks are sampled from the validation split during validation and the test split during testing. Sampled tasks from mini-ImageNet have fixed ways and shots. The Meta-Dataset benchmark tasks have variable ways and shots, which are randomly sampled. We use the same implementation and sampler settings as defined for the Meta-Dataset benchmark~\cite{metadataset}.

As described in Section \ref{sec:validation}, we experiment with two types of few-shot classifiers (Linear and N-Centroid). If a linear classifier is used, a new weight matrix must be learned for mapping the features to the target classes. We train the linear classifier for $75$ steps using SGD with momentum of $0.9$. The learning rate and $L_2$ regularization factor is fixed to $0.01$ and  $0.001$, respectively, in all experiments.

We ran BOHB on $30$-$40$ parallel GPU (Nvidia P100) workers using the default settings
of $\eta=3$ and the number of mini-batch updates as a budget parameter (see Section~\ref{sec:bohb}). 
Each worker trains a feature extractor and computes the validation accuracy on few-shot tasks. 
For mini-ImageNet and Meta-Dataset, we allocate a maximum training budget of $480k$ and $720k$ mini-batch updates respectively. These numbers were chosen assuming a batch size of $16$. Note that we also have the batch size as a hyperparameter in our search spaces. Therefore, the number of updates need to be scaled up or down (for the same effective number of epochs). For instance if the sampled batch sizes are $8$ or $32$ for mini-ImageNet, the resulting mini-batch updates will be $960k$ and $240k$, respectively. We ran BOHB for $35$ successive halving iterations. 

\textbf{Search space ranges.} The search space ranges for operations in search spaces $S1$ and $S2$ are shown in Table~\ref{t:search_range}. 

\begin{table}[h]
\begin{center}
	\caption{Search space ranges} 
\label{tab:search_space_range}
\resizebox{0.60\linewidth}{!}{
\begin{tabular}{ll}
\hline
\textbf{Hyperparameter}    &  \textbf{Range/Values} \\
\hline
Optimization params & \\ 
\hline 
Optimizer                 &  \{SGD, ADAM\} \\
Learning rate             &   $[1e^{-4}, 0.5e^{-1}]$ \\
L2 regularization         &  $[1e^{-5}, 5e{-3}]$ \\
Decay every               &  $[100,200 \cdots 100,000]$ \\
Batch size                &  $\{2^2, 2^3 \cdots 2^6\}$ \\
\hline
Augmentation params & \\ 
\hline 
Number of operations ($N_{ops}$)   & $[1,10]$ \\ 
Standard deviation for $op_{i}$ ($\sigma_{i}$) & $[1,25]$ \\ 
\hline 

\end{tabular}

}
\label{t:search_range}
\end{center}
\end{table}

\subsection{Impact of hyperparameter optimization}

\begin{table}[h]
\begin{center}
	\caption{HPO on mini-ImageNet. Test accuracies are reported for $5$-way $1$-shot and $5$-shot tasks. } 
\label{tab:hpo_mini}
\resizebox{0.75\linewidth}{!}{
\begin{tabular}{llcc}
\hline
\textbf{Classifier}                  &   \textbf{Hyperparams}                  &   \multicolumn{2}{c}{\textbf{Test accuracy}}                  \\
                                     &                                             &    1-shot                  &   5-shot         \\
\hline
Linear                       &  Random                             &   $47.52\pm0.73$                &   $70.46\pm0.60$            \\
Linear           			 &  Default                            &   $47.96\pm 0.72$               &   $72.12\pm0.63$            \\
Linear                       &  BOHB                     		   &   $54.55\pm0.80$                &   $76.99\pm0.62$            \\
\hline
N-Centroid                   & Random                              &   $51.34\pm0.77$                &   $72.22\pm0.62$            \\
N-Centroid                   & Default                             &   $50.37\pm0.79$                &   $74.21\pm0.66$            \\
N-Centroid                   & BOHB                 			   &   $\mathbf{59.64\pm0.77}$        &   $\mathbf{78.29\pm0.59}$   \\
\hline
\end{tabular}


}
\end{center}
\end{table}

\begin{table}[h]
\begin{center}
	\caption{HPO on the Meta-Dataset. We report average test accuracies for each dataset group. $D1$ shows within-domain, and $D2+D3$ shows cross-domain performance (see Table \ref{tab:dspool}). 
}
\label{tab:hpo_meta}
\resizebox{0.75\linewidth}{!}{
\begin{tabular}{llcc}
\hline
\textbf{Classifier}          &   \textbf{Hyperparams}                  &  \multicolumn{2}{c}{\textbf{Test Accuracy}}      \\
& & $(D1)$ & ($D2 + D3)$\\
\hline
Linear                       &  Random                                             &   $36.35\pm0.95$                       &  $42.11\pm0.97$   \\
Linear           			 &  Default                                            &   $22.78\pm0.83$                       &  $26.77\pm1.04$            \\
Linear                       &  BOHB                     		                   &   $50.60\pm1.15$            &   $57.98\pm1.10$              \\
\hline
N-Centroid                       & Random                                          &   $46.52\pm1.02$            &  $55.26\pm0.98$    \\
N-Centroid                       & Default                                         &   $46.72\pm1.11$            &  $56.34\pm1.06$ \\
N-Centroid                       & BOHB                 			               &   $\mathbf{51.92\pm1.05}$   &  $\mathbf{59.96\pm0.95}$\\
\hline
\end{tabular}


%


}
\end{center}
\end{table}

To assess the impact of hyperparameters on few-shot performance we performed an experiment of training the feature extractor under three settings: 1) with randomly sampled hyperparameters; 2) with hyperparameters provided by publicly available code (result of manual optimization); 3)~with hyperparameters obtained with BOHB. For sampling hyperparameters in 1) and 3) we used the same search space $S1$, consisting of optimization hyperparameters only. For the random baseline we report the  mean and standard deviation averaged over $15$ runs. For each setting, we allocated the same training budget. In the default setting, we used hyperparameter choices from Chen \etal~\cite{closerlook}. 
When running BOHB on the Meta-dataset, we used $D1$ for training and validation, and we performed separate HPO for each classifier type (Linear or N-centroid). 

Table~\ref{tab:hpo_mini} and Table~\ref{tab:hpo_meta} show results on mini-ImageNet and Meta-dataset, respectively. 
In both cases, there is a clear benefit of HPO over random and default parameters. 
The N-centroid classifier always outperformed the linear classifier, even for the Meta-dataset dataset, where test datasets may have very different data distribution ($D2$ and $D3$) compared to the training and validation set. 
Thus, for the remaining experiments we only report the results with the N-centroid classifier.

\subsection{Optimizing for different data distributions}

\begin{table}[h]
\begin{center}
\caption{Test performance of models optimized with BOHB (search space $S1$) and ImageNet-GBM for training. Cross-domain validation improves the domain transfer compared to within-domain validation even if the validation set is in a different domain than the test set.}
\label{tab:better_transfer}
\resizebox{0.9\linewidth}{!}{

\begin{tabular}{lcccc}
\hline 
\multicolumn{1}{l}{\bf Validation}  &  \multicolumn{4}{c}{\textbf{Test Accuracy}}            \\ 
                                                                    &  \multicolumn{1}{c}{ ImageNet-GBM}  &  \multicolumn{1}{c}{ Omniglot}  &  \multicolumn{1}{c}{ Quickdraw}  &  \multicolumn{1}{c}{ Birds} \\ \hline 
 ImageNet-GBM            &   $51.92\pm1.05$              &   $67.57\pm1.21$              &   $50.33\pm1.04$              &   $70.69\pm0.90$            \\ 
                                                                   Omniglot                    &   $49.60\pm1.10$              &   $\mathbf{72.03\pm1.24}$     &   $53.81\pm1.08$              &   $68.51\pm0.95$            \\ 
                                                                   QuickDraw                   &   $50.56\pm1.10$              &   $67.86\pm1.29$              &   $\mathbf{54.50\pm1.04}$     &   $69.87\pm0.91$            \\ 
                                                                   Birds                       &   $\mathbf{52.08\pm1.08}$     &   $68.43\pm1.18$              &   $51.87\pm1.03$              &   $\mathbf{72.10\pm0.93}$            \\ 
\hline 
\end{tabular}
}
\end{center}
\end{table}

Can HPO help learn a feature extractor from ImageNet which transfers better to target domains (under a limited data regime)? 
We trained on ImageNet-GBM's \emph{train split} and validated on few-shot tasks sampled from a different domain's \emph{validation split}. We consider QuickDraw (consisting of sketch images) and Omniglot (handwritten character images) as domains that are very different from ImageNet. As similar domain, we consider the Birds dataset. 
After optimization, the feature extractor is evaluated on few-shot tasks sampled from a target domain's \emph{test split}. We only use the search space $S1$ for this experiment. The results are shown in Table~\ref{tab:better_transfer}. 

\textbf{Improved transfer to other domains.} Using the target domain's data for validation is useful to make a feature extractor that transfers better to that domain. For QuickDraw and Omniglot, this leads to an improvement by $4\%$-$5\%$ over an optimized baseline which uses ImageNet as a the validation domain. 
For the Birds dataset, we see only a small improvement, because the validation domain is already mostly included in the training domain. 

It is worth noting that optimizing for a target domain (e.g. Omniglot) does not destroy much of the performance on other domains (e.g. ImageNet). Thus, there is no over-fitting problem as if one would finetune the network on Omniglot, which would destroy its performance on ImageNet.

\textbf{Transferring to mixed domains.} We also experimented with a validation objective, where few-shot tasks are sampled from a \emph{mixture} of different domains. We used the validation splits from the dataset group $D2$; see Table \ref{tab:dspool}. The results are shown in Table \ref{tab:mixed_transfer}. We still observe a performance gain but the effect is diminished compared to validating on single datasets. 

\begin{table}[h]
\begin{center}
	\caption{Mixed domain transfer. Feature extractors are trained on $D1$ and validated on few-shot tasks sampled from $D2$ (see Table \ref{tab:dspool}). We report average test performance on both $D1$ and $D2$. }
\label{tab:mixed_transfer}
\resizebox{0.65\linewidth}{!}{

\begin{tabular}{ccc}
\hline
 \textbf{Validation} &  \multicolumn{2}{c}{\textbf{Test Accuracy}} \\
			                 & $D1$             & $D2$                \\ 
\hline
$D1$       & $\mathbf{51.92\pm1.05}$   & $62.82\pm0.93$      \\
$D2$       & $50.75\pm1.10$   & $\mathbf{63.73\pm0.94}$      \\
\hline\end{tabular}

}
\end{center}
\end{table}

\subsection{Evaluating generalization on unseen domains}

In the previous section we showed that it is possible to tune a feature extractor, such that it performs better on target domains that are different from the training domain by using a validation objective from that domain.

It would be even more practical, if we could learn universal features that generalize to domains unseen during training or validation. To this end, we performed a cross-validation experiment with feature extractors trained on ImageNet-GBM and validated on $C_{v}$ and tested on $C_t$, where, $C_{v}$ and $C_{t}$ are disjoint set of datasets which do not include ImageNet-GBM. Few-shot tasks are sampled uniformly at random from $C_v$ (during validation)  and $C_t$ (during testing). 

\begin{table}[t]
\caption{Cross-validation splits for evaluating generalization. Cross validation splits are: $C1$, $C2$, $C3$, $C4$.} 
\centering
\label{tab:cross_splits}
\resizebox{0.65\linewidth}{!}{ 
\begin{tabular}{lll}
\hline 
\textbf{Split} & \textbf{Validation}  & \textbf{Test}  \\
 & ($C_v$) &  ($C_t$) \\
\hline 
$C1$                  &    \makecell[l]{Aircraft, Birds, Textures, \\ Quickdraw, Fungi}           &     \makecell[l]{ VGG Flower, Omniglot, \\ Traffic Sign, MSCOCO } \\ \hline 
$C2$                  &    \makecell[l]{VGG Flower, Omniglot, Fungi, \\ Aircraft, QuickDraw}     &      \makecell[l]{ Birds,  Textures, \\ Traffic Sign, MSCOCO } \\   \hline 
$C3$                  &    \makecell[l]{Birds, Textures, Omniglot,  \\ VGG Flower, Fungi }        &     \makecell[l]{ Aircraft, QuickDraw, \\ Traffic Sign, MSCOCO } \\  \hline 
$C4$                  &    \makecell[l]{QuickDraw, Birds, Textures,  \\ VGG Flower, Aircraft }     &     \makecell[l]{ Fungi,  Omniglot, \\ Traffic Sign, MSCOCO }  \\ 
\hline 
\end{tabular}

}
\end{table}

We construct four cross-validation splits ($C1$, $C2$, $C3$, $C4$) each having validation and test subset (shown in Table \ref{tab:cross_splits}). Since Traffic Sign and MSCOCO are always used only for testing in the Meta-Dataset framework~\cite{metadataset}, we do not use them for validation. For each cross-validation split, we run BOHB to optimize the features using its validation subset. The best model is then tested on the unseen datasets in the split. The test results are shown in Table~\ref{tab:crossvalidataion}. To compute a final generalization score, for each dataset we take the average across non-zero test scores across each cross-validation split. 
Finally, we take an average over accuracy and standard deviations (in the last column) across each dataset. We compare the average performance with a model validated on ImageNet-GBM only (last row). 

\textbf{Consistent generalization.} Results in Table \ref{tab:crossvalidataion} show that the cross-validation score is similar to that of a model trained and validated on ImageNet-GBM only. Irrespective of the validation dataset used, we do not observe a tendency to overfit and our models generalize equally well on unseen domains. However, we also do not achieve a substantial benefit regarding generalization to unseen domains from using cross-validation: the average performance is similar to validation only on ImageNet-GBM.

\begin{table}[t]
\begin{center}
\caption{Cross validation results. We report results for each test dataset in a cross validation split. All validation datasets in a split are marked with "-". On the rightmost column we report average scores for each dataset. The final cross validation score is computed by taking the average across this column. 
}\label{tab:crossvalidataion}
\resizebox{0.9\linewidth}{!}{ 

\begin{tabular}{lccccc}
\hline 
\multicolumn{1}{l}{\bf Test source}  &  \multicolumn{1}{c}{ $\mathbf{C1}$}  &  \multicolumn{1}{c}{$\mathbf{C2}$}  &  \multicolumn{1}{c}{$\mathbf{C3}$}  &  \multicolumn{1}{c}{$\mathbf{C4}$}  &  \multicolumn{1}{c}{\bf Dataset avg.}  \\ 
\hline 
Omniglot                    &   $68.86\pm1.20$              &   $-$                         &   $-$                         &   $67.30\pm1.21$              &   $68.08\pm1.20$              \\ 
Aircraft                    &   $-$                         &   $-$                         &   $55.76\pm0.93$              &   $-$                         &   $55.76\pm0.93$              \\ 
Birds                       &   $-$                         &   $68.12\pm0.98$              &   $-$                         &   $-$                         &   $68.12\pm0.98$              \\ 
Textures                    &   $-$                         &   $68.50\pm0.68$              &   $-$                         &   $-$                         &   $68.50\pm0.68$              \\ 
QuickDraw                   &   $-$                         &   $-$                         &   $50.93\pm1.02$              &   $-$                         &   $50.93\pm1.02$              \\ 
Fungi                       &   $-$                         &   $-$                         &   $-$                         &   $41.16\pm1.09$              &   $41.16\pm1.09$              \\ 
VGG Flower                  &   $87.37\pm0.59$              &   $-$                         &   $-$                         &   $-$                         &   $87.37\pm0.59$              \\ 
Traffic Sign                &   $55.68\pm1.06$              &   $52.67\pm1.06$              &   $57.92\pm1.06$              &   $53.97\pm1.00$              &   $55.06\pm1.04$              \\ 
MSCOCO                      &   $46.97\pm1.03$              &   $46.65\pm1.08$              &   $46.61\pm1.00$              &   $47.51\pm1.05$              &   $46.96\pm1.04$              \\ 
\hline 
\multirow{2}{*}{Average acc.}& \multicolumn{4}{c}{After cross validation }               &   $60.22\pm0.95$              \\ 
&\multicolumn{4}{c}{Model validated with ImageNet-GBM}                             &   $59.96\pm0.95$              \\ 
\hline 
\end{tabular}
}
\end{center}
\end{table}

\subsection{Optimization of data augmentation parameters}

\begin{figure}[h]
\begin{center}
\includegraphics[width=0.8\linewidth]{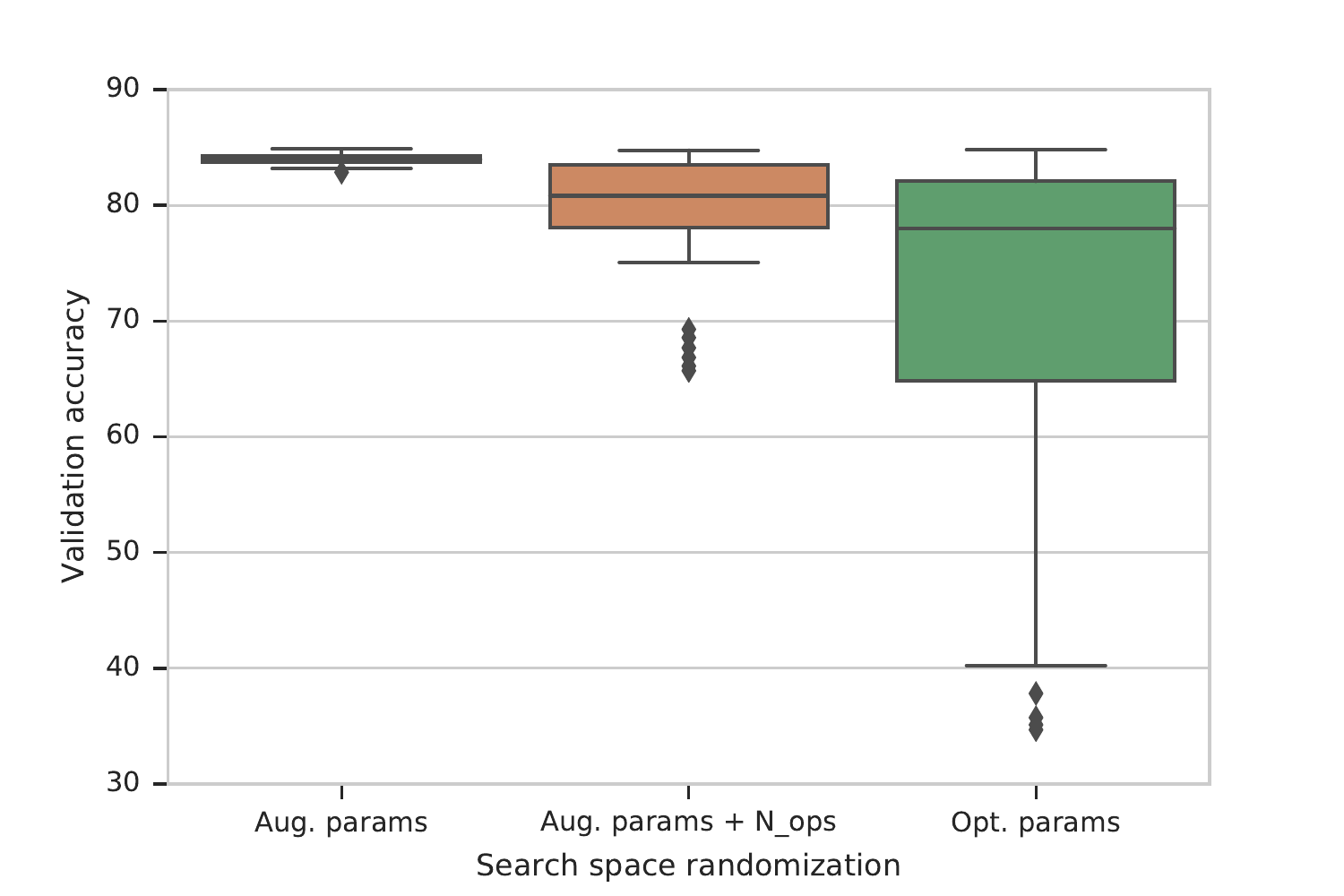}
\caption{Performance variation with search space randomization. Randomizing the optimization search space shows biggest performance variation, suggesting that they are sensitive and thus more important to tune. }
\label{fig:searchspace_rnd}
\end{center}
\end{figure}

We pick-up the questions from Section~\ref{sec:search_space} and test if HPO can be useful in the context of data augmentation or if it is superfluous. We used the larger search space $S2$ described in Section~\ref{sec:search_space}. 
We experiment with three settings: First, tuning optimization hyperparameters and standard deviation ($\sigma_{i}$)  for each operation and $N_{ops}$ (search space $S2$). Second, modifying $S2$ to have a common standard deviation ($\sigma_{common}$) for each operation and thirdly, tuning optimization hyperparameters with the condition that augmentation operations have random $\sigma$'s and the number of operations $N_{ops}\!=\!1$. 
The results are reported in Table~\ref{tab:aug} and show that adding data augmentation to the search space \emph{does} improve performance. The improvement is not as large as the gain from optimizing the parameters in S1, but it is substantial. 
We observed that the validation objective is more sensitive towards hyperparameters which are more important. To analyze, we train multiple models by fixing one subset and randomizing the rest from $S2$. We consider three cases: 1) freeze optimization parameters
and randomize $\sigma_i$ 2)  randomize $N_{ops}$ along with  $\sigma_i$ 3) freeze augmentation and randomize optimization parameters. The variance of the validation objective is shown in Figure~\ref{fig:searchspace_rnd}. We observe that randomizing the optimization hyperparameters show the largest performance variance, which indicates that they must be carefully tuned.

\begin{table}[h]
\begin{center}
\caption{Tuning data augmentation. We report test performance on mini-ImageNet. The first and second row represent models trained with fixed augmentation settings (random crops, flipping and intensity changes from~\cite{closerlook}). In the second row we tune the optimization parameters (search space $S1$). The third row shows performance after tuning search space $S2$. In the fourth row, we tune a common standard deviation $\sigma_{common}$ per augmentation operation. The configuration in the last row fixes $N_{ops}$ to $1$ and samples random $\sigma_i$ for each operation. The first column denotes if the optimization parameters were tuned or not. 
\label{tab:aug}
}

\resizebox{0.9\columnwidth}{!}{




\begin{tabular}{clcc}
\hline
\textbf{Tune Opt.} & \textbf{Data aug.}                          &   \multicolumn{2}{c}{\textbf{Test accuracy}}  \\

                     &                                           &    1-shot                  &   5-shot         \\
\hline 
\xmark               & default (fixed)                              &    $50.37\pm0.79$                  &  $74.21\pm0.66$ \\        
\cmark               & default (fixed)                                   &   $59.63\pm0.64$           &   $78.44\pm0.45$     \\
\cmark               & $\{ N_{ops}, \sigma_{1} \cdots \sigma_{10}\} $    &   $\mathbf{62.12\pm0.62}$           &   $\mathbf{81.03\pm0.45}$         \\
\cmark               &  $\{N_{ops},\sigma_{common} \} $                  &   $59.79\pm0.78$                      &   $80.27\pm0.55$         \\
\cmark               & $N_{ops}=1$, $\{\sigma_{1} \cdots \sigma_{10}\}_{rand}$ &   $59.43\pm0.63$                 &   $80.36\pm0.58$        \\
\hline
\end{tabular}

}
\end{center}
\end{table}

\subsection{Comparison to the state of the art}

With the improved generalization obtained with HPO, how do the results compare to the state of the art in few-shot learning? 
In Table \ref{tab:mini_imagenet_sota} we observe that BOHB-N-Centroid tuned on the simple search space $S1$ (optimization only) with fixed standard augmentation (random crops, flipping and intensity changes) clearly improves over comparable methods~\cite{closerlook}. Using the larger search space $S2$ (augmentation \& optimization), we are comparable to the test accuracy of a $20$ network ensemble~\cite{ensembles}. We also report our results after ensembling.  Unlike~\cite{ensembles} we do not retrain the best model multiple times, but rather take the ensemble from the top $20$ models discovered by BOHB. 

Also on Meta-Dataset, results  compare favorably to existing approaches (see Table~\ref{tab:metadataset_sota}). It is worth noting that our performance gains come without the need for any additional adaptation (i.e the feature extractor does not need to be optimized on few-shot tasks during evaluation). 
Since, ensembling provides such large performance gains~\cite{ensembles}, we also report ensemble results for the Meta-dataset benchmark. Since the compared methods do not use ensembles, this comparison is not on the same ground anymore, yet shows that the gains are mostly complementary.  Also,  we found that optimizing hyperparameters using the larger search space $S2$ did not lead to an improvement with ImageNet as training source. We conjecture that ImageNet data provides sufficient regularization to our ResNet18 feature extractor and additional data augmentation does not help. 

\begin{table}[t]
\begin{center}
\caption{Comparison to state-of-the-art (mini-ImageNet).  We report results using only the nearest centroid classifier (BOHB-NC). The suffixes $S1$ and $S2$ denote the search space used. Since different methods use different input image resolutions and network backbones we also indicate that. The column "Ensm." determines whether a model uses ensembling during evaluation. 
}
\label{tab:mini_imagenet_sota}

\resizebox{\columnwidth}{!}{

\begin{tabular}{lllcccc}
\hline 
\textbf{Model}   & \textbf{Aug.}          & \textbf{Network} & \textbf{Input}    & \textbf{Ensm.}  &  \multicolumn{2}{c}{\textbf{Test accuracy}}\\ 
             & & & &                                                                      &  1-shot & 5-shot \\ \hline  \\ 
TADAM      ~\cite{tadam} & standard       & ResNet12             & $84$                   &     \xmark     &   $58.50\pm0.30$         &  $76.70\pm0.30$                \\ 
MetaOptNet ~\cite{metaoptnet}& standard     & ResNet12             & $84$                    &     \xmark     &   $62.64\pm0.61$         &  $78.63\pm0.46$               \\ 
LEO        ~\cite{leo}  & standard       & WideResNet           & $80$                    &     \xmark     &  $61.76\pm0.08$         &  $77.59\pm0.12$                 \\ 
FEAT       ~\cite{feat} & standard       & WideResNet           & $80$                    &     \xmark     &  $61.72\pm0.11$         &  $78.32 \pm 0.16$          \\ 
Robust20   ~\cite{ensembles}  & standard  & WideResNet           & $80$                    &     \cmark     &   $63.46\pm0.62$         &  $81.94\pm0.44$               \\
Robust20   ~\cite{ensembles}  & standard  & ResNet18             & $224$                   &     \cmark     &   $63.95\pm0.61$         &  $81.59\pm0.42$              \\
Linear     ~\cite{closerlook} & standard & ResNet18             & $224$                   &     \xmark     &   $51.75\pm0.80$         &  $74.27\pm0.63$                \\
Cosine     ~\cite{closerlook}  & standard & ResNet18             & $224$                  &     \xmark     &   $51.87\pm0.77$         &  $75.68\pm0.63$                \\
\hline 
BOHB-NC-$S1$               & standard  & ResNet18             &  $224$                 &     \xmark     & $59.63\pm0.64$          & $78.44\pm0.45$ \\
               & standard  & ResNet18             &  $224$                 &     \cmark     & $66.09\pm0.66$          & $83.35\pm0.41$ \\
\hline 
BOHB-NC-$S2$               & learned   & ResNet18             &  $224$                 &     \xmark     & $62.12\pm0.62$          & $81.03\pm0.45$ \\
                & learned   & ResNet18             &  $224$                 &     \cmark     & $\mathbf{66.41\pm0.62}$ & $\mathbf{84.52\pm0.41}$ \\
\hline 
\end{tabular}
}
\end{center}
\end{table}

\begin{table}[t]
\begin{center}
\caption{Comparison to state-of-the-art (Meta-Dataset). We compare performance of our models (BOHB-NC-S1) to other approaches on the Meta-Dataset benchmark (numbers are from ~\cite{metadataset}). We report results of models trained on $D1$ and optimized within-domain ($D1$) and cross-domain ($D2$) validation sets. We also report performance after ensembling top $20$ models in each case (denoted by ensm.). The "Adapt" column indicates if parameters of the feature extractor are also optimized during few-shot evaluation. Note, here the baseline methods also use hyperparameter optimization~\cite{metadataset}. 
}
\label{tab:metadataset_sota}

\resizebox{0.9\columnwidth}{!}{
\begin{tabular}{lcccccc}
\hline 
\multicolumn{1}{l}{\bf Method} &  \multicolumn{1}{c}{\bf Val}&  \multicolumn{1}{c}{\bf Adapt}  &  \multicolumn{3}{c}{\textbf{Test accuracy}} \\ 
 &   &                                                            &  \multicolumn{1}{c}{$D1$}  &  \multicolumn{1}{c}{$D2$}  &  \multicolumn{1}{c}{$D3$} \\ \hline \\ 
Finetune            &  $D1$      &   \cmark                      &   $45.78\pm1.10$              &   $60.32\pm1.12$              &   $50.83\pm1.14$            \\ 
MAML                &  $D1$      &   \cmark                      &   $36.09\pm1.01$              &   $42.28\pm1.07$              &   $30.38\pm1.23$            \\ 
Proto-MAML          &  $D1$     &   \cmark                      &   $51.01\pm1.05$              &   $61.64\pm1.01$               &   $47.30\pm1.08$            \\ 
k-NN                &  $D1$      &   \xmark                      &   $41.03\pm1.01$              &   $50.24\pm0.94$              &   $37.48\pm1.09$            \\ 
MatchingNet         &  $D1$      &   \xmark                      &   $45.00\pm1.10$              &   $54.94\pm0.97$              &   $41.39\pm1.07$            \\ 
RelationNet         &  $D1$      &   \xmark                      &   $34.69\pm1.01$              &   $47.31\pm0.98$              &   $31.41\pm1.03$            \\ 
ProtoNet            &  $D1$      &   \xmark                      &   $50.50\pm1.08$              &   $60.34\pm1.02$              &   $44.06\pm1.10$            \\ 
\hline 
BOHB-NC-S1     &  $D1$      &   \xmark                      &   $51.92\pm1.05$        &   $62.82\pm0.93$              &   $49.91\pm1.02$            \\ 
               &  $D2$      &   \xmark                      &   $50.75\pm1.10$                 &   $63.73\pm0.94$     &   $50.39\pm1.02$            \\ 
\hline 
BOHB-NC-S1 (ensm.)     &  $D1$      &   \xmark                      &   $\mathbf{55.39\pm1.06}$      &   $68.72\pm0.87$              &   $54.69\pm1.01$            \\ 
                       &  $D2$      &   \xmark                      &   $54.47\pm1.07$      &   $\mathbf{68.85\pm0.89}$              &   $\mathbf{54.70\pm1.02}$            \\ 
\hline 
\end{tabular}


}
\end{center}
\end{table}

\section{Analysis}

\begin{figure}[t]
\begin{center}
\subcaptionbox{BOHB-N-Centroid (mini-ImageNet) \label{fig:pp_dist_meta}}{\includegraphics[width=\linewidth]{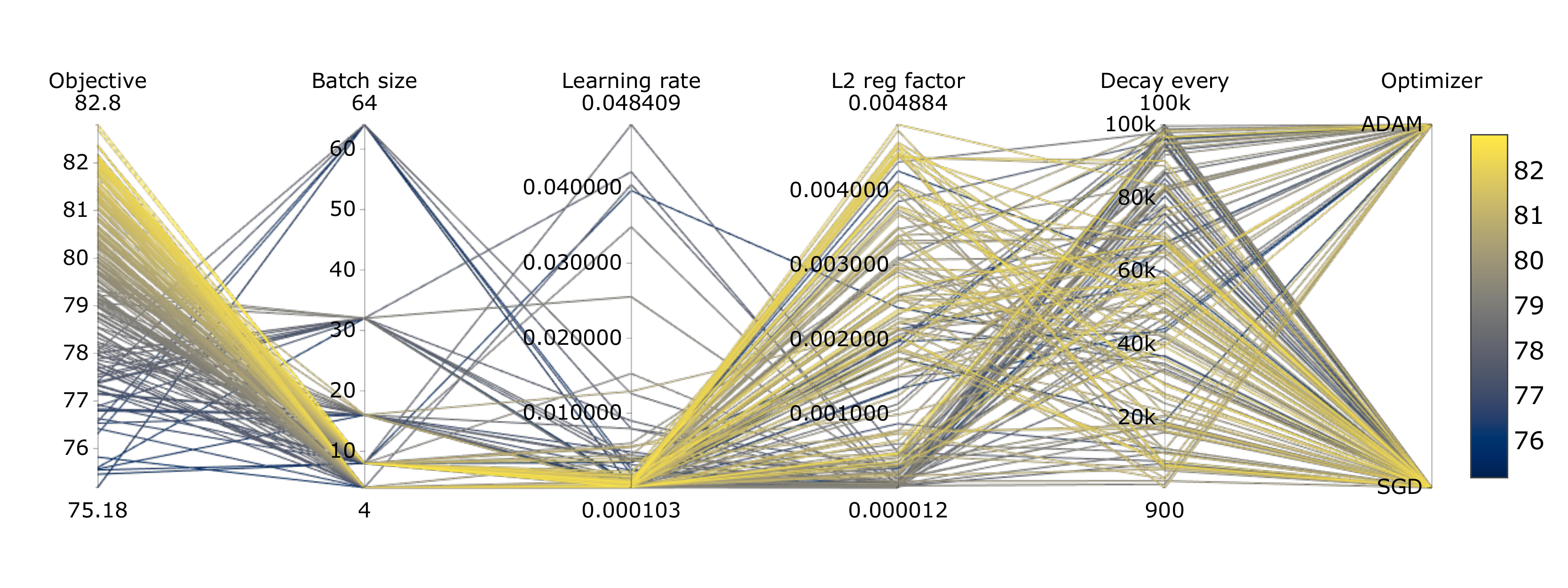}}%
\\
\subcaptionbox{BOHB-N-Centroid (ImageNet-GBM) \label{fig:pp_dist_meta}}{\includegraphics[width=\linewidth]{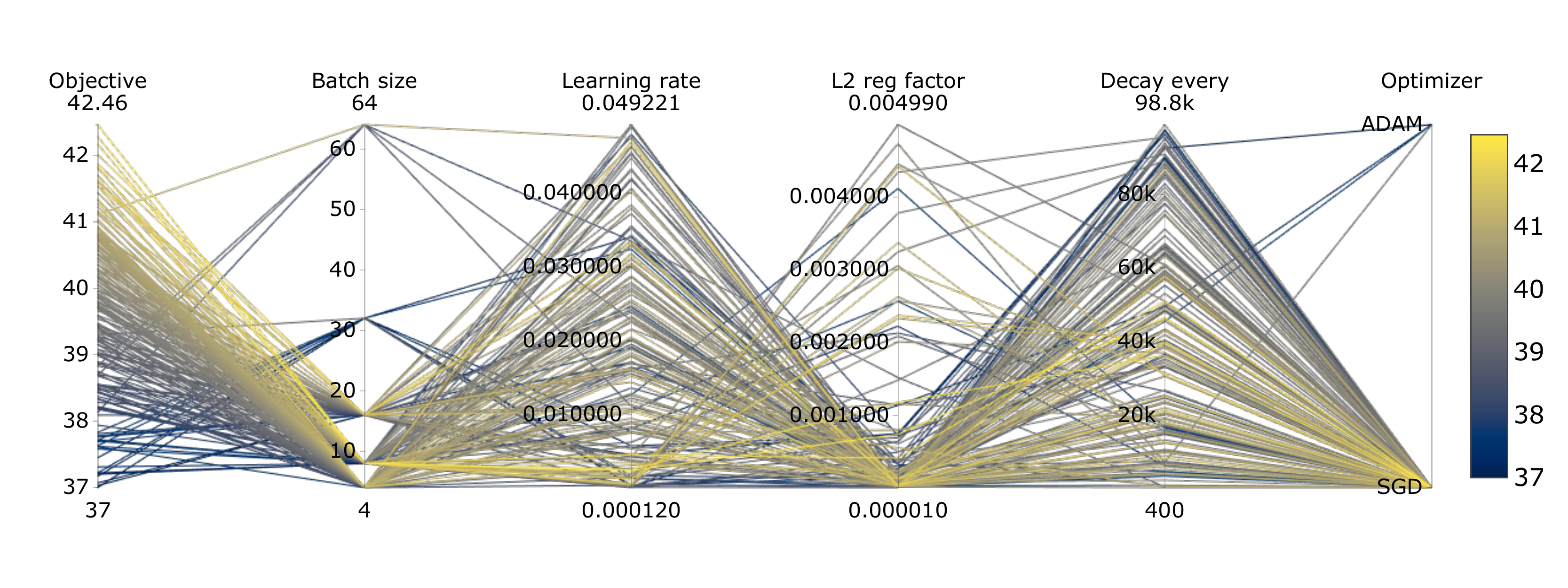}}%
\caption{Hyperparameter relationships. Parallel coordinate plots are shown for BOHB runs using  N-Centroid classifiers on mini-ImageNet (top) and ImageNet-GBM (bottom). The first axis represents the validation objective (accuracy). We  consider samples in the good performance region, i.e, within $(V_{max}\!-\!5, V_{max})$, where $V_{max}$  represents the maximum validation accuracy. The remaining axes represent hyperparameters in search space $S1$. The numbers above and below each axis shows the minimum and maximum value. A connected line across the axes map a validation accuracy to a hyperparameter configuration. Yellow and blue correspond to good and bad hyperparameter configurations.}
\label{fig:parallel_plots}
\end{center}

\end{figure}

\textbf{What are good hyperparameters?} We use parallel coordinate plots (PCP)~\cite{pcp} to visualize relationships between the validation objective and the different hyperparameters in our search space (shown in Figure~\ref{fig:parallel_plots}).
We observe that configurations with  lower learning rate and smaller batch size (resulting in more mini-batch updates) typically lead to better performance. 
This is in line with the common understanding in the community~\cite{small_batch_training}. In practice, we found that a batch size of $8$ consistently performed well. 
For mini-ImageNet experiments, BOHB finds good configurations with both ADAM and SGD, however for experiments on ImageNet-GBM, SGD clearly outperforms ADAM. 
With regard to  search space $S2$ we observed that the number of randomly sampled augmentation operations ($N_{ops}$) is often chosen to be one or two, i.e., it is advantageous to apply transformations one after the other. More details are in the supplementary material.

 \section{Conclusions}
 In this paper we aimed for features that generalize well across tasks and across domains. We showed that hyperparameter optimization with few-shot classification as validation objective is a very powerful tool to this end. Apart from the typical within-domain analysis, we also investigated few-shot learning across domains. We saw that HPO adds to cross-domain generalization even when the optimization is not run across domains but on a rich enough dataset like ImageNet. Moreover, it provides a way to adapt to a specific domain without destroying generalization on other domains. We shedded more light on the ongoing discussion whether data augmentation can benefit from optimizing its parameters and got a positive answer for a reasonably sized search space. Finally, we found that HPO is well compatible with ensembles.

 {\small
 \bibliographystyle{ieee_fullname}
 \bibliography{egbib}
 }

\newcommand{\beginsupplement}{%
        \setcounter{table}{0}
        \renewcommand{\thetable}{(\roman{table})}%
        \setcounter{figure}{0}
        \renewcommand{\thefigure}{(\roman{figure})}%
        \setcounter{section}{0}
     }

\newpage
\clearpage

\beginsupplement

\twocolumn[{\centering{ \Huge Supplementary Material}\vspace{3ex} \medbreak \vspace{2ex}}]

\appendix

\section{Data augmentation}

\subsection{Number of augmentation operations}
\begin{figure}[h]
 \begin{center}
 \includegraphics[width=0.9\linewidth]{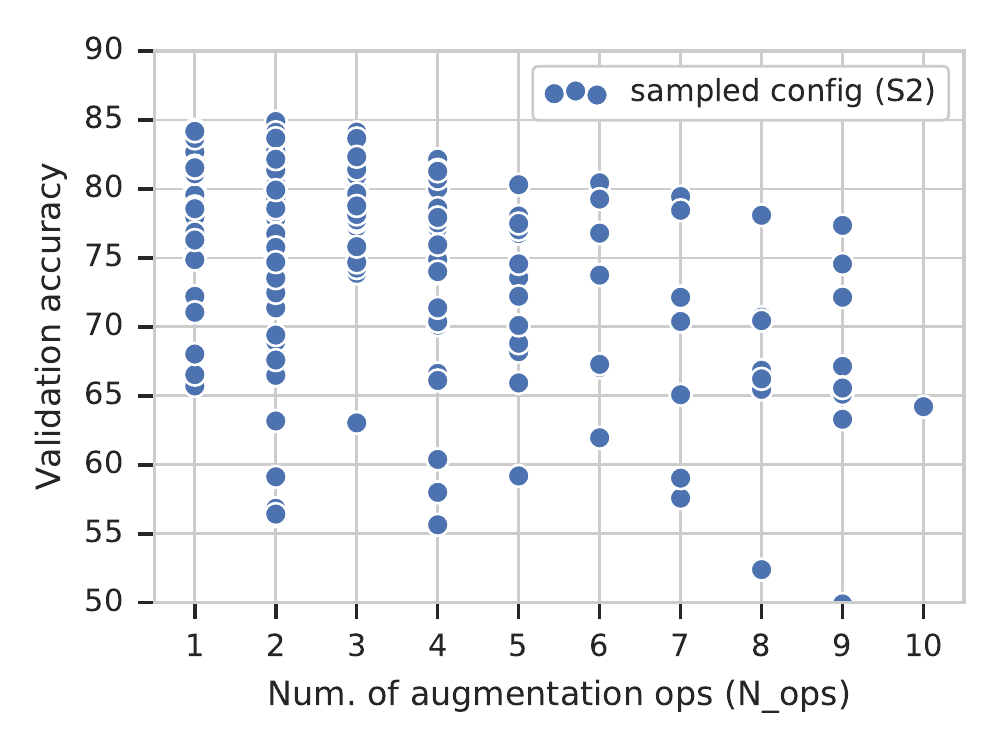}
 \caption{Optimal values for $N_{ops}$. We plot the validation accuracy vs number of augmentation operations per mini-batch ($N_{ops}$) for a BOHB run on mini-ImageNet while optimizing hyperparameters on search space $S2$. Each circle represents a configuration sampled by BOHB.}
 \label{fig:num_ops}
 \end{center}
 \end{figure}

During our experiments with BOHB on search space $S2$ (see Section 3.1), we found that good configurations had $N_{ops}$  (the number of data augmentation operations applied per mini-batch) set to low values, as shown in Figure~\ref{fig:num_ops}. The best configuration had $N_{ops}\!=\!2$. We observe that as $N_{ops}$ increases, the validation accuracy drops.

\subsection{Results for Meta-Dataset on search space S2}
We found that optimizing hyperparameters using the larger search space $S2$ (optimization + augmentation) did not lead to an improvement on the Meta-Dataset benchmark. The results compared to search space $S1$ are shown in Table~\ref{tab:metadataset_s2}. We conjecture that training data from ImageNet-GBM ($D1$) provides sufficient regularization to our ResNet18 feature extractor and any additional data augmentation does not help.
\begin{table}[t]
\begin{center}
\caption{Results on Meta-Dataset (with search spaces  $S1$ and $S2$). We compare test performance on the three dataset groups (described in Section 4.1 of the paper). The BOHB models were trained and validated on $D1$. The suffixes $S1$ and $S2$ denote the search space used.
}
\label{tab:metadataset_s2}

\resizebox{0.8\columnwidth}{!}{
\begin{tabular}{lcccc}
\hline 
\multicolumn{1}{l}{\bf Method}  &  \multicolumn{3}{c}{\textbf{Test accuracy}} \\ 
                                                            &  \multicolumn{1}{c}{$D1$}  &  \multicolumn{1}{c}{$D2$}  &  \multicolumn{1}{c}{$D3$} \\ \hline \\ 
BOHB-NC-S1                         &   $51.92\pm1.05$                 &   $62.82\pm0.93$              &   $49.91\pm1.02$            \\ 
BOHB-NC-S2                        &   $50.75\pm1.09$                 &   $61.67\pm0.95$           &   $48.27\pm1.00$            \\ 
\hline 
\end{tabular}


}
\end{center}
\end{table}

\section{More results with a linear classifier}

\subsection{Cross-domain validation}

We report results using the linear classifier with cross-domain validation, again showing better transfer on tasks from a different data distribution (see Section 4.3). The results are shown in Table~\ref{tab:better_transfer_linear} and are complementary to Table 5 in the main paper (where a N-Centroid classifier was used).

\begin{table}[h]
\begin{center}
\caption{Test performance of models trained on ImageNet-GBM and optimized with BOHB (search space $S1$). Cross-domain validation improves the domain transfer. Here, we report results using the Linear classifier for few shot classification tasks. These results are complementary to Table 5 in the main paper.}
\label{tab:better_transfer_linear}
\resizebox{\linewidth}{!}{
\begin{tabular}{lcccc}
\hline 
\multicolumn{1}{l}{\bf Validation}  &  \multicolumn{4}{c}{\textbf{Test Accuracy}}            \\ 
                                                                    &  \multicolumn{1}{c}{ ImageNet-GBM}  &  \multicolumn{1}{c}{ Omniglot}  &  \multicolumn{1}{c}{ Quickdraw}  &  \multicolumn{1}{c}{ Birds} \\ \hline 
   ImageNet-GBM                &   $\mathbf{50.60\pm1.15}$              &   $64.09\pm1.36$              &   $46.26\pm1.27$              &   $67.68\pm1.08$            \\ 
                                                                         Omniglot                    &   $46.93\pm1.20$              &   $\mathbf{66.25\pm1.38}$              &   $42.10\pm1.39$         &   $63.79\pm1.12$            \\ 
                                                                         QuickDraw                   &   $44.59\pm1.09$              &   $64.89\pm1.33$                       &   $\mathbf{51.15\pm1.11}$     &   $61.26\pm1.09$            \\ 
                                                                         Birds                       &   $49.10\pm1.14$              &   $66.23\pm1.31$                       &   $50.04\pm1.22$  & $\mathbf{69.77\pm1.00}$\\  
\hline 
\end{tabular}
}
\end{center}
\end{table}

\subsection{Hyperparameter relationships}

\begin{figure}[h]
\begin{center}
\subcaptionbox{BOHB-Linear (mini-ImageNet)     \label{fig:pp_linear_mini}}{\includegraphics[width=\linewidth]{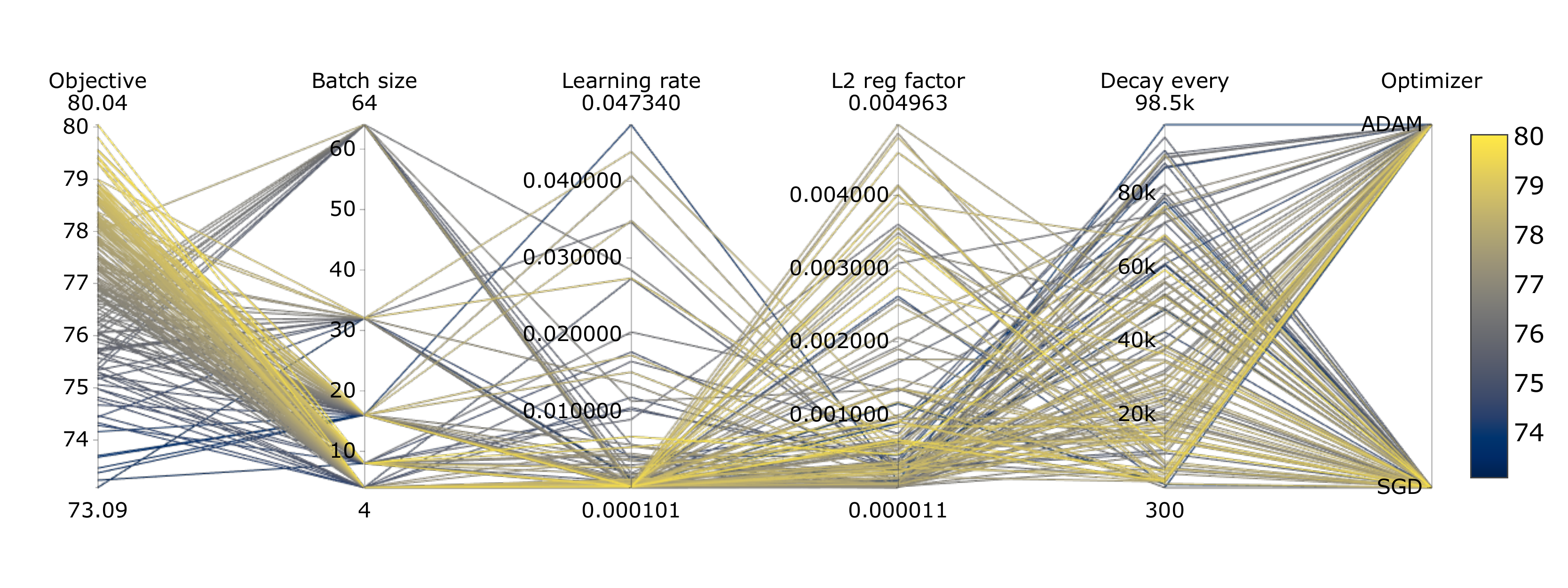}}%
\hfill
\\
\subcaptionbox{BOHB-Linear (ImageNet-GBM)     \label{fig:pp_linear_mini}}{\includegraphics[width=\linewidth]{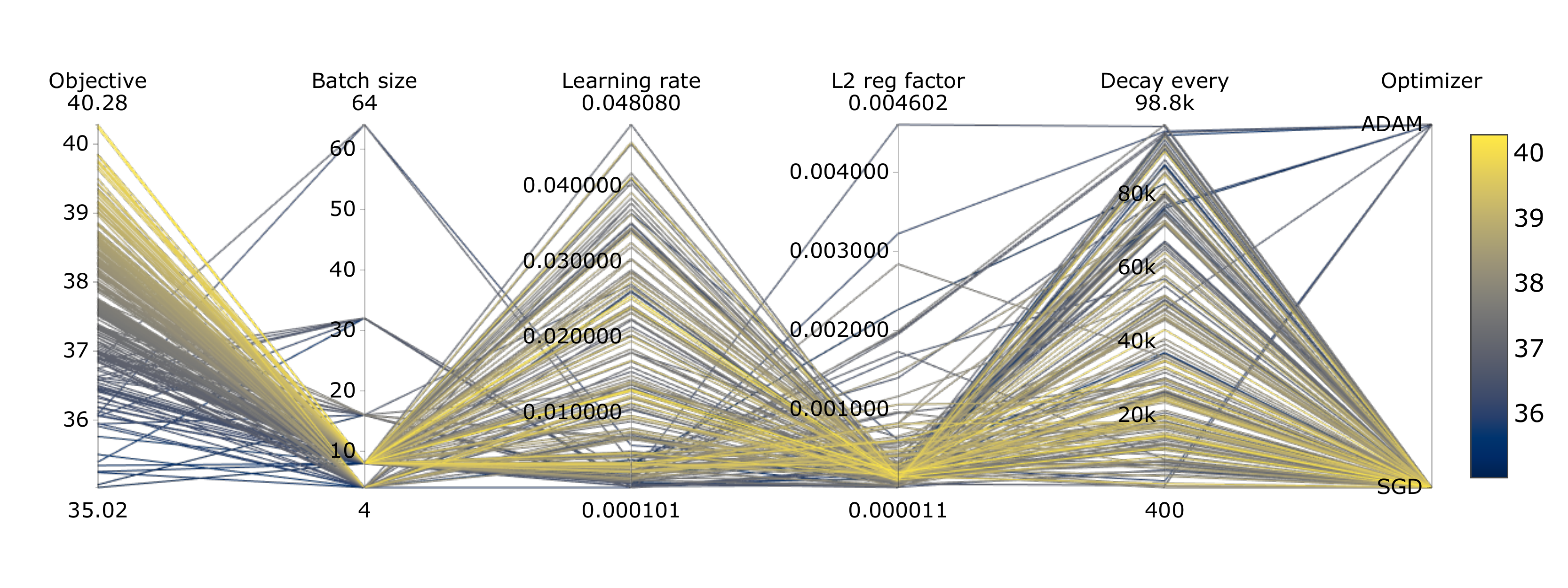}}%
\caption{Hyperparameter relationships. Parallel coordinate plots are shown for BOHB runs using  Linear classifiers on mini-ImageNet (top) and ImageNet-GBM (bottom). The first axis represents the validation objective (accuracy). The remaining axes represent hyperparameters in search space $S1$. }
\label{fig:parallel_plots_linear}
\end{center}

\end{figure}

Similar to Figure 3 in the main paper, we use parallel coordinate plots to visualize relationships between the validation objective and the different hyperparameters in our search space, while using the linear classifier (shown in Figure~\ref{fig:parallel_plots_linear}).
Supporting the discussion in Section 5, we show that configurations with smaller batch sizes lead to optimal performance, also with the linear classifier. A similar pattern is observed for the choice of optimizer (SGD outperforms ADAM on ImagNet-GBM).

\section{Ensembling}
\begin{figure}[t]
\begin{center}
\subcaptionbox{1-shot test performance  \label{fig:ens_1shot}}{\includegraphics[width=\linewidth]{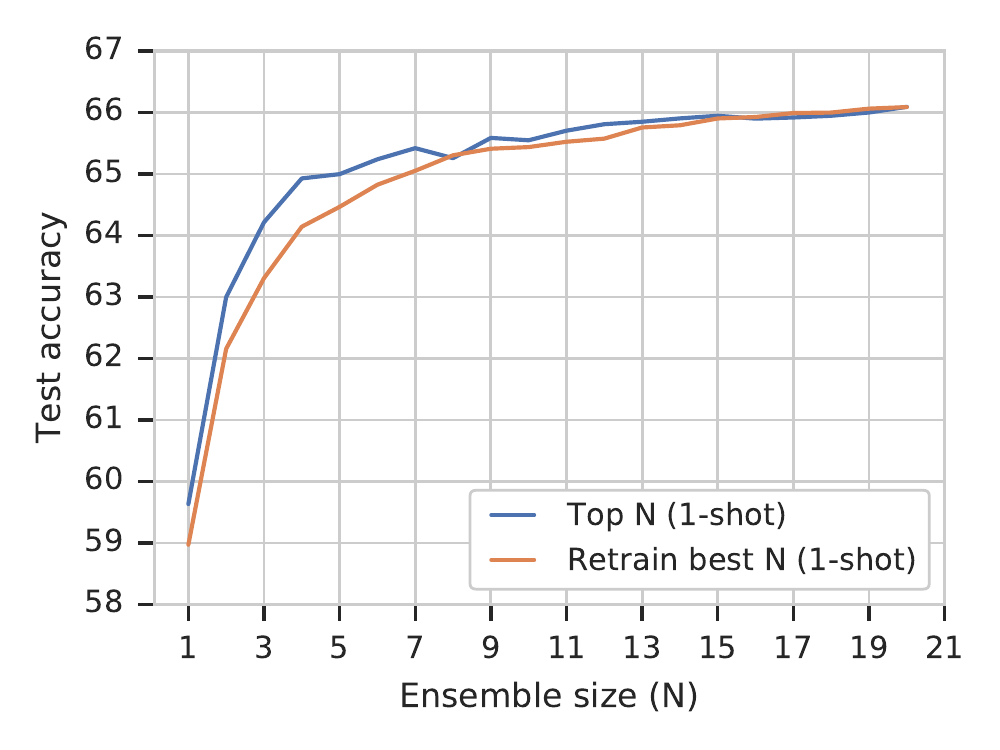}}%

\subcaptionbox{5-shot test performance \label{fig:ens_5shot}}{\includegraphics[width=\linewidth]{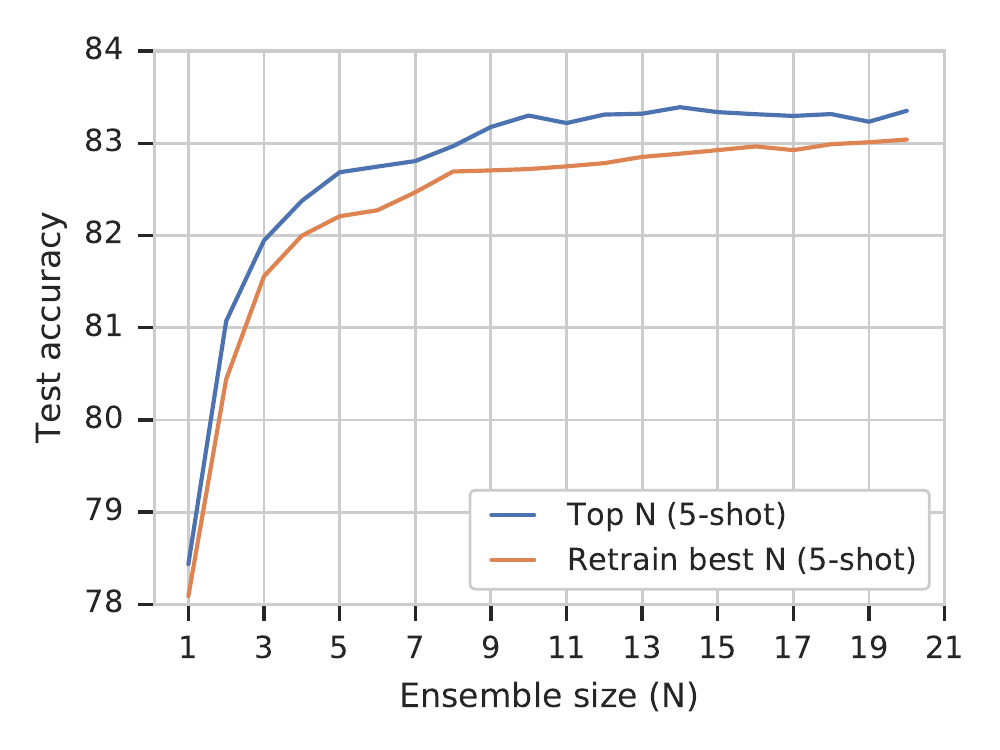}}%
\caption{Performance vs ensemble size. We visualize the effect of increasing the ensemble size on test accuracy for 5-way few shot tasks sampled from mini-ImageNet. The top and bottom figures show performance trajectories for 1-shot and 5-shot tasks respectively. The blue curve denotes the performance trajectory of ensembling the top $N$ configurations from BOHB, while the orange curve shows the one obtained by re-training the best model (found by BOHB) $N$ times. }

\label{fig:ensembling}
\end{center}
\end{figure}

In Figure~\ref{fig:ensembling}, we observe that ensembling logits from top $N$ models obtained from BOHB perform better on 5-shot tasks compared to an ensemble of the same size obtained by re-training the best model. However, on 1-shot tasks, we end up having the same performance in both cases as the ensemble size increases.

\section{Detailed results on Meta-Dataset}
We report more detailed results showing performance on individual datasets from Meta-Dataset. Results for all our models in comparison to baselines from the Meta-Dataset benchmark are shown in Table~\ref{tab:results_metadataset_full}.

\begin{figure*}[h]
\begin{center}
\includegraphics[width=\linewidth]{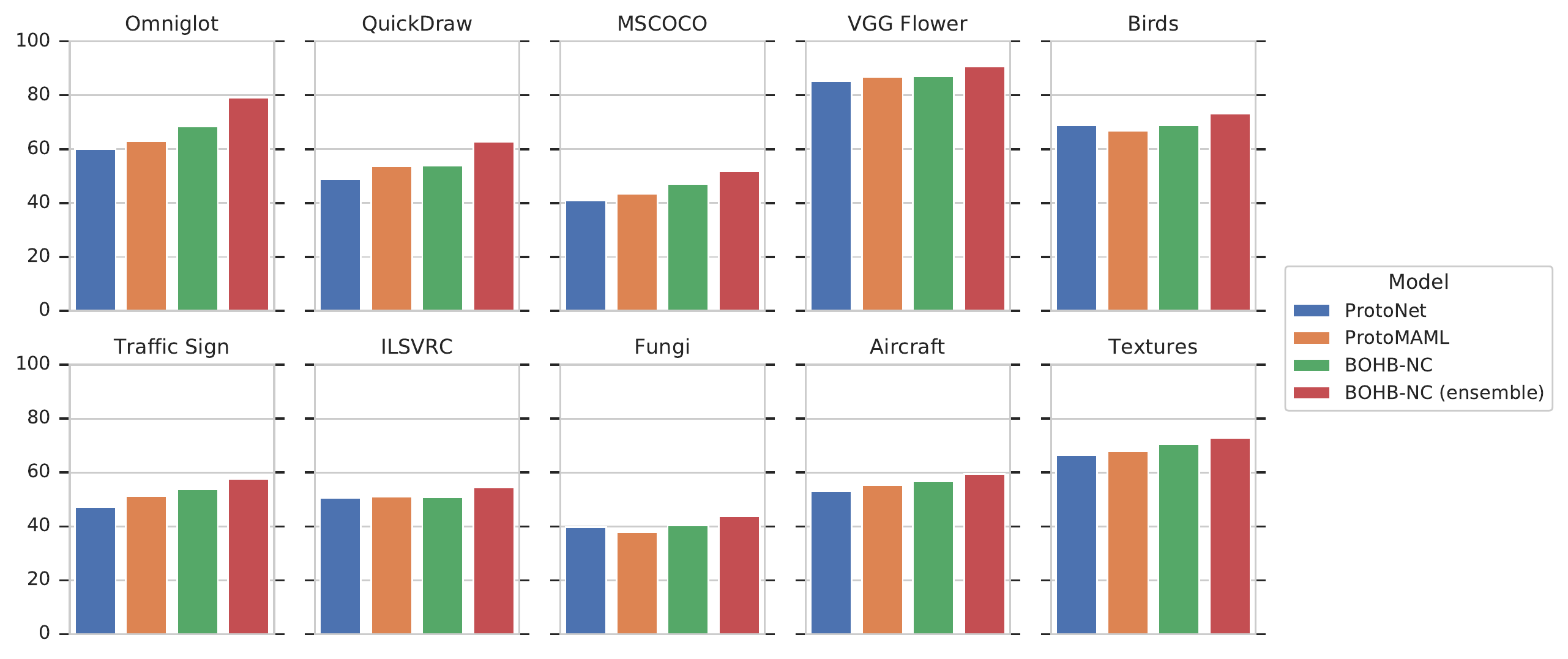}
\caption{Visualizing performance on individual datasets from Meta-Dataset. Each model uses Imagenet-GBM as training source. In comparision to baselines ProtoNet and ProtoMAML, BOHB-NC has a performance which is comparable or better. With ensembling we can see large boosts in performance, even on datasets from a different data distributions (example; QuickDraw and Omniglot). }
\label{fig:compare}
\end{center}

\end{figure*}

A visual comparison of our best models with baseline models (ProtoNet and ProtoMAML) is shown in Figure~\ref{fig:compare}. We observe that BOHB-NC has a performance which is comparable or better. With ensembling we can see large boosts in performance, even for difficult datasets such as QuickDraw and Omniglot which are from a very different data distribution.

\begin{table*}[h]
\begin{center}
\caption{Results on Meta-Dataset. We report results of our models (prefixed with BOHB) on all datasets from the Meta-Dataset benchmark. The suffixes "L", "NC" indicate the choice of few shot classifier; linear or nearest centroid. "NC-E" indicates that the results are from an ensembled model using the nearest centroid classifier. The validation dataset(s) used is shown under the model name. We compare to top 2 baselines from Meta-Dataset benchmark with adaption (Finetune, Proto-MAML) and without adaption (ProtoNet, MatchingNet). Here adaption indicates if the parameters of the feature extractor were optimized during few shot evaluation.
}
\label{tab:results_metadataset_full}
\resizebox{\textwidth}{!}{

\begin{tabular}{lcccccccccc}
\hline

  &  \multicolumn{2}{c}{\emph{With adaption}}   & \multicolumn{2}{c}{\emph{No adaption}} & \multicolumn{6}{c}{\emph{No adaption (Ours)}}\\ \cmidrule(lr){2-3}\cmidrule(lr){4-5} \cmidrule(lr){6-11}
\bf Test source  &  \bf Finetune  &  \bf Proto-MAML &
    \bf MatchingNet  &  \bf ProtoNet  &  \bf BOHB-L  &
    \bf BOHB-NC  & \bf BOHB-L  &  \bf BOHB-NC  &
    \bf BOHB-NC-E  &  \bf BOHB-NC-E  \\
    & Val. $D1$ & Val. $D1$ & Val. $D1$ & Val. $D1$ & Val. $D1$& Val. $D1$ & Val. $D2$ & Val. $D2$ 
    & Val. $D1$ & Val. $D2$\\\hline \\
ILSVRC                      &   $45.78\pm1.10$              &   $51.01\pm1.05$              &   $45.00\pm1.10$              &   $50.50\pm1.08$              &   $50.60\pm1.15$              &   $51.92\pm1.05$              &   $50.40\pm1.11$              &   $50.75\pm1.10$              &   $\mathbf{55.39\pm1.06}$              &   $54.47\pm1.07$            \\ 
Omniglot                    &   $60.85\pm1.58$              &   $63.00\pm1.35$              &   $52.27\pm1.28$              &   $59.98\pm1.35$              &   $64.09\pm1.36$              &   $67.57\pm1.21$              &   $64.61\pm1.38$              &   $68.45\pm1.26$              &   $77.45\pm1.06$              &   $\mathbf{79.12\pm1.06}$            \\ 
Aircraft                    &   $68.69\pm1.26$              &   $55.31\pm0.96$              &   $48.97\pm0.93$              &   $53.10\pm1.00$              &   $57.36\pm1.04$              &   $54.12\pm0.90$              &   $57.97\pm1.06$              &   $56.75\pm0.92$              &   $\mathbf{60.85\pm0.91}$              &   $59.48\pm0.99$            \\ 
Birds                       &   $57.31\pm1.26$              &   $66.87\pm1.04$              &   $62.21\pm0.95$              &   $68.79\pm1.01$              &   $67.68\pm1.08$              &   $70.69\pm0.90$              &   $68.54\pm1.00$              &   $68.93\pm0.95$              &   $\mathbf{73.56\pm0.83}$              &   $73.17\pm0.91$            \\ 
Textures                    &   $69.05\pm0.90$              &   $67.75\pm0.78$              &   $64.15\pm0.85$              &   $66.56\pm0.83$              &   $70.38\pm0.91$              &   $68.34\pm0.76$              &   $68.43\pm0.93$              &   $70.56\pm0.72$              &   $\mathbf{72.86\pm0.71}$              &   $72.81\pm0.65$            \\ 
QuickDraw                   &   $42.60\pm1.17$              &   $53.70\pm1.06$              &   $42.87\pm1.09$              &   $48.96\pm1.08$              &   $46.26\pm1.27$              &   $50.33\pm1.04$              &   $48.61\pm1.25$              &   $54.00\pm1.05$              &   $61.16\pm0.93$              &   $\mathbf{62.76\pm0.94}$            \\ 
Fungi                       &   $38.20\pm1.02$              &   $37.97\pm1.11$              &   $33.97\pm1.00$              &   $39.71\pm1.11$              &   $33.82\pm1.03$              &   $41.38\pm1.12$              &   $37.32\pm1.15$              &   $40.43\pm1.06$              &   $\mathbf{44.54\pm1.08}$     &   $43.89\pm1.15$            \\ 
VGG Flower                  &   $85.51\pm0.68$              &   $86.86\pm0.75$              &   $80.13\pm0.71$              &   $85.27\pm0.77$              &   $85.51\pm0.75$              &   $87.34\pm0.59$              &   $85.95\pm0.73$              &   $87.02\pm0.65$              &   $90.62\pm0.55$              &   $\mathbf{90.73\pm0.50}$            \\ 
Traffic Sign                &   $66.79\pm1.31$              &   $51.19\pm1.11$              &   $47.80\pm1.14$              &   $47.12\pm1.10$              &   $55.17\pm1.33$              &   $51.80\pm1.04$              &   $55.93\pm1.29$              &   $53.69\pm1.04$              &   $57.53\pm1.02$              &   $\mathbf{57.61\pm1.00}$            \\ 
MSCOCO                      &   $34.86\pm0.97$              &   $43.41\pm1.06$              &   $34.99\pm1.00$              &   $41.00\pm1.10$              &   $41.58\pm1.17$              &   $48.03\pm0.99$              &   $41.65\pm1.11$              &   $47.09\pm0.99$              &   $\mathbf{51.86\pm0.99}$              &   $51.79\pm1.03$            \\ 
\hline 
$D1$                      &   $45.78\pm1.10$              &   $51.01\pm1.05$              &   $45.00\pm1.10$              &   $50.50\pm1.08$              &   $50.60\pm1.15$              &   $51.92\pm1.05$              &   $50.40\pm1.11$              &   $50.75\pm1.10$              &   $\mathbf{55.39\pm1.06}$              &   $54.47\pm1.07$            \\ 

$D2$                          &   $60.32\pm1.12$              &   $61.64\pm1.01$              &   $54.94\pm0.97$              &   $60.34\pm1.02$              &   $60.73\pm1.06$              &   $62.82\pm0.93$              &   $61.63\pm1.07$                &   $63.73\pm0.94$              &   $68.72\pm0.87$              &   $\mathbf{68.85\pm0.89}$            \\ 
$D3$                          &   $50.83\pm1.14$              &   $47.30\pm1.08$              &   $41.39\pm1.07$              &   $44.06\pm1.10$              &   $48.38\pm1.25$              &   $49.91\pm1.02$              &   $48.79\pm1.20$                &   $50.39\pm1.02$              &   $54.69\pm1.01$              &   $\mathbf{54.70\pm1.02}$            \\ 
\hline 
\end{tabular}

}

\end{center}
\end{table*}

\end{document}